\documentclass[a4paper,twoside]{article}

\usepackage{epsfig}
\usepackage{subcaption}
\usepackage{calc}
\usepackage{amssymb}
\usepackage{amstext}
\usepackage{amsmath}
\usepackage{amsthm}
\usepackage{multicol}
\usepackage{pslatex}

\usepackage{apalike}
\let\bibhang\relax
\usepackage{natbib}

\usepackage{tabularx}
\usepackage{makecell}
\usepackage{multicol}
\usepackage{graphicx}
\usepackage{tikz}
\usetikzlibrary{arrows, positioning, fit}
\usepackage{pgfplots}
\usepgfplotslibrary{external}

\usepackage{xspace}
\usepackage{xcolor}
\usepackage[hyphens]{url}
\usepackage{footnote}

\graphicspath{{images/}}
\usepackage{xcolor}

\definecolor{cadmiumgreen}{rgb}{0.0, 0.42, 0.24}
\definecolor{cornellred}{rgb}{0.7, 0.11, 0.11}
\definecolor{cornflowerblue}{rgb}{0.39, 0.58, 0.93}
\definecolor{darkgreen}{RGB}{  0,100, 0} 
\definecolor{firebrick}{RGB}{178, 34,34}
\definecolor{gray}{rgb}{0.80, 0.80, 0.80}
\definecolor{hlcolor}{rgb}{1.00, 1.00, 0.5}

\newcommand{\figref}[1]{Fig.~\ref{#1}}                         
\newcommand{\tabref}[1]{Table~\ref{#1}}                        

	\makeatletter
	\DeclareRobustCommand\onedot{\futurelet\@let@token\@onedot}
	\def\@onedot{\ifx\@let@token.\else.\null\fi\xspace}                
	\def\eg{\emph{e.g}\onedot} 			
	             
	\def\ie{\emph{i.e}\onedot} 			
	             
	\def\wrt{w.r.t\onedot} 				 
	                    
	\def\etal{\emph{et al}\@onedot}          

	\makeatother

\usepackage{SCITEPRESS}     

\begin{document}


\title{A Photogrammetry-based Framework to Facilitate Image-based Modeling and Automatic Camera Tracking}

\author{\authorname{Sebastian Bullinger\sup{1}\orcidAuthor{0000-0002-1584-5319}, Christoph Bodensteiner\sup{1}\orcidAuthor{0000-0002-5563-3484} and Michael Arens\sup{1}\orcidAuthor{0000-0002-7857-0332}}
\affiliation{\sup{1}Department of Object Recognition, Fraunhofer IOSB, Ettlingen, Germany}
\email{\{sebastian.bullinger, christoph.bodensteiner, michael.arens\}@iosb.fraunhofer.de}
}

\abstract{We propose a framework that extends Blender to exploit Structure from Motion (SfM) and Multi-View Stereo (MVS) techniques for image-based modeling tasks such as sculpting or camera and motion tracking. Applying SfM allows us to determine camera motions without manually defining feature tracks or calibrating the cameras used to capture the image data. With MVS we are able to automatically compute dense scene models, which is not feasible with the built-in tools of Blender. Currently, our framework supports several state-of-the-art SfM and MVS pipelines. The modular system design enables us to integrate further approaches without additional effort. The framework is publicly available as an open source software package.}
\onecolumn \maketitle \normalsize \setcounter{footnote}{0} \vfill

%

\section{\uppercase{Introduction}}
\label{sec:introduction}

%
%
%
%
%
%
%
%
%
%
\subsection{Photogrammetry-based Modeling and Camera Tracking}
\noindent Many tasks in the area of image-based modeling or visual effects such as sculpting or motion and camera tracking involve a substantial amount of user interaction to achieve satisfying results. Even with many modern tools like Blender \citep{Blender2020}, designers require to perform many steps manually. With the recent progress of \emph{Structure from Motion} (SfM), \emph{Multi-View Stereo} (MVS) and texturing techniques the automation of specific steps such as the determination of the camera motion and the reconstruction of the scene geometry (including texture computation) has become feasible. \\
Using common mesh data formats, the majority of modeling tools allows to import the reconstructed geometry and corresponding textures of state-of-the-art MVS \citep{FuhrmannToG2014,JancosekISRN2014,Schoenberger2016mvs,Ummenhofer2017} and texturing libraries \citep{BurtToG1983,WaechterECCV2014}. However, such data formats do not include camera calibration and camera motion information, which is crucial for many tasks such as creating visual effects.
\begin{figure}[t]
	\centering{\includegraphics[width=0.47\textwidth]{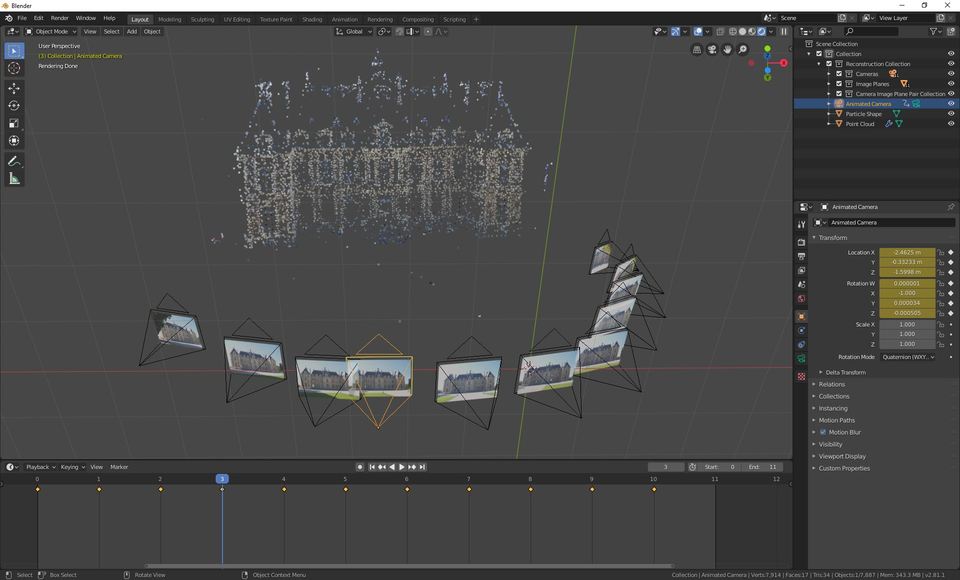}}
	\caption{Reconstruction result of the sceaux dataset \citep{Moulon2012Sceaux} in the 3D view of Blender including the reconstructed camera poses as well as the corresponding point cloud.}
	\label{fig_imported_result}
\end{figure}  \\
\newlength{\widthOverviewStaticSfMInput}
\setlength{\widthOverviewStaticSfMInput}{0.08\textwidth}
\newlength{\widthOverviewStaticSfMRec}
\setlength{\widthOverviewStaticSfMRec}{3.0\widthOverviewStaticSfMInput}
\newcommand\firstrowStaticSfM{0}
\newcommand\firstrowStaticSfMShifted{0.01\textwidth}
\newcommand\firstcolStaticSfM{0}
\newcommand\secondcolStaticSfM{0.13\textwidth}
\newcommand\thirdcolStaticSfM{0.28\textwidth}
\newcommand\fourthcolStaticSfM{0.58\textwidth}
\newcommand\fifthcolStaticSfM{0.8\textwidth}
\newlength{\blockRoundedTextWidth}
\setlength{\blockRoundedTextWidth}{1.5cm}
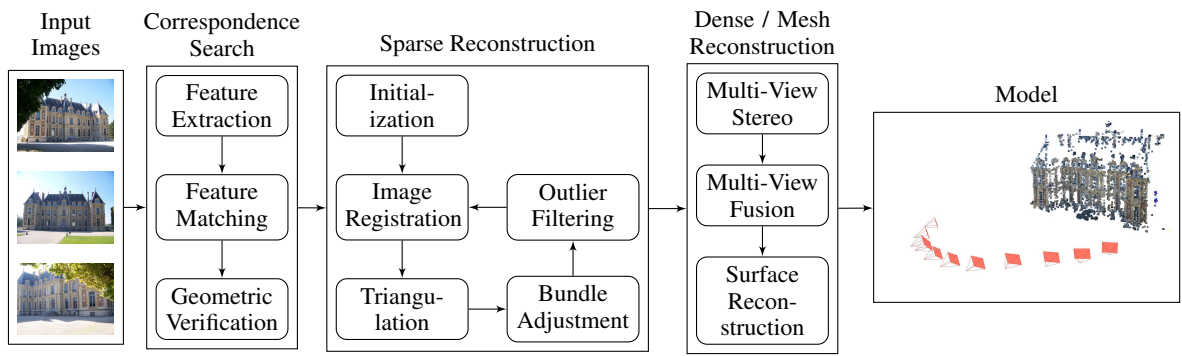
\begin{figure*}[t]
	\centering

	\tikzsetnextfilename{figure_structure_from_motion_pipeline}
	\begin{tikzpicture}[auto]

		\newcommand{\figureTextSize}{\footnotesize}
		
		
		\tikzstyle{blockRounded} = [
		rectangle, 
		draw, 
		text width=\blockRoundedTextWidth, 
		text centered, 
		rounded corners, 
		font=\figureTextSize
		]
		
		\tikzstyle{blockCorner} = [
		rectangle, 
		draw, 
		text centered, 
		font=\figureTextSize
		]
		
		\tikzstyle{line} = [
		draw, 
		-latex', 
		font=\figureTextSize, 
		text centered
		]
		
		\tikzstyle{inputImageStyle} = [
		inner sep=0,
		text width = \widthOverviewStaticSfMInput
		]
		
		\tikzstyle{recImageStyle} = [
		draw,
		text width = \widthOverviewStaticSfMRec
		]


		\begin{scope}[every path/.style=line]

		\node[inputImageStyle] (second_input) at (\firstcolStaticSfM,\firstrowStaticSfM) {
			\includegraphics[ width = \widthOverviewStaticSfMInput]{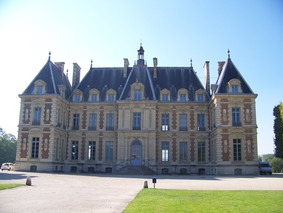}
		};

		\node[inputImageStyle, above = 0.25 cm of second_input] (first_input)  {
			\includegraphics[ width = \widthOverviewStaticSfMInput]{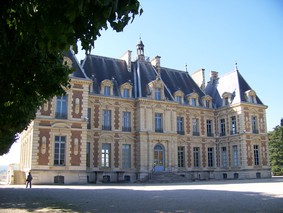}
		};
	
		\node[inputImageStyle, below = 0.25 cm of second_input] (third_input) {
			\includegraphics[ width = \widthOverviewStaticSfMInput]{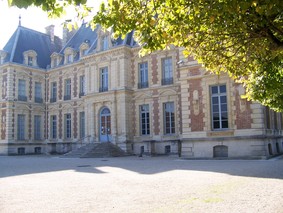}
		};
		\node[blockCorner,label={[text width=\widthOverviewStaticSfMInput] Input Images}, fit=(first_input)(second_input)(third_input), draw] (input_images) {};


		\node [blockRounded, draw] (feature_matching) at (\secondcolStaticSfM,\firstrowStaticSfM) {Feature \\ Matching};
		\node [blockRounded, above = 0.5 cm of feature_matching, draw] (feature_extraction) {Feature \\ Extraction}; 
		\node [blockRounded, below = 0.5 cm of feature_matching, draw] (geometric_verification) {Geometric Verification};  
		
		\draw (feature_extraction) -- (feature_matching);
		\draw (feature_matching) -- (geometric_verification);
		
		\node[blockCorner,label={[text width=0.15\textwidth] Correspondence Search}, fit=(feature_extraction)(feature_matching)(geometric_verification), draw] (correspondence_search) {};
		
		\draw (input_images) -- (correspondence_search);
		
		
		
		\node [blockRounded, draw] (image_registration) at (\thirdcolStaticSfM,\firstrowStaticSfM) {Image \\ Registration};
		\node [blockRounded, above = 0.5 cm of image_registration, draw] (initialization) {Initial- \\ization}; 
		\node [blockRounded, below = 0.5 cm of image_registration, draw] (triangulation) {Triangu- \\lation}; 
		
		\node [blockRounded, right = 0.5 cm of image_registration, draw] (outlier_filtering) {Outlier  Filtering};  
		\node [blockRounded, right = 0.5 cm of triangulation, draw] (bundle_adjustment) {Bundle \\ Adjustment}; 
		
		\draw (initialization) -- (image_registration);
		\draw (image_registration) -- (triangulation);
		\draw (triangulation) -- (bundle_adjustment);
		\draw (bundle_adjustment) -- (outlier_filtering);
		\draw (outlier_filtering) -- (image_registration);
		
		\node[blockCorner,label={Sparse Reconstruction}, fit=(initialization)(image_registration)(triangulation)(outlier_filtering)(bundle_adjustment), draw] (sparse_reconstruction) {};
		
		\draw (correspondence_search) -- (sparse_reconstruction);


		\node [blockRounded, draw] (multi_view_fusion) at (\fourthcolStaticSfM,\firstrowStaticSfMShifted) {Multi-View Fusion};
		\node [blockRounded, above = 0.4 cm of multi_view_fusion, draw] (multi_view_stereo) {Multi-View Stereo}; 
		\node [blockRounded, below = 0.4 cm of multi_view_fusion, draw] (surface_reconstruction) {Surface \\ Recon- \\struction};  
		
		\draw (multi_view_stereo) -- (multi_view_fusion);
		\draw (multi_view_fusion) -- (surface_reconstruction);
		
		\node[blockCorner,label={[text width=0.15\textwidth] Dense / Mesh \\ Reconstruction}, fit=(multi_view_stereo)(multi_view_fusion)(surface_reconstruction), draw] (dense_reconstruction) {};
		
		\draw (sparse_reconstruction) -- (dense_reconstruction);
		
		\node[recImageStyle, label={Model}] (output) at (\fifthcolStaticSfM,\firstrowStaticSfM) {
			\includegraphics[ width = \widthOverviewStaticSfMRec]{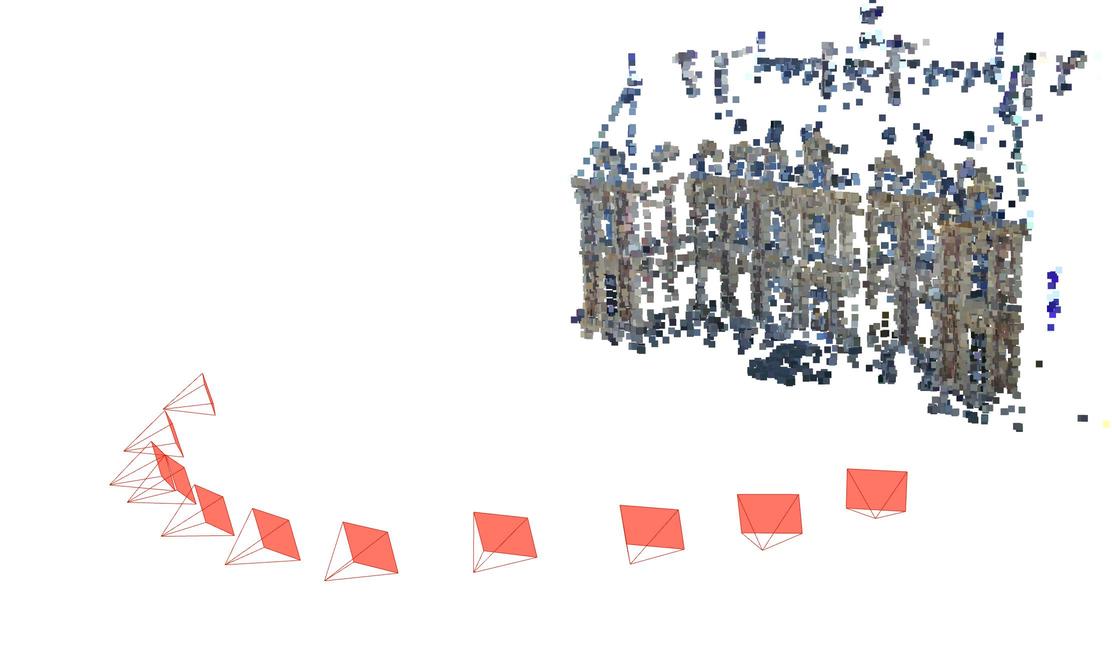}
		};
		
		\draw (dense_reconstruction) -- (output);
		
		\end{scope}	

		\end{tikzpicture}
	
	\caption{Building blocks of state-of-the-art incremental Structure from Motion and Multi-view Stereo pipelines. The input images are part of the Sceaux Castle dataset \cite{Moulon2012Sceaux}.}
	
	\label{figure_structure_from_motion_pipeline}
\end{figure*}

In order to overcome these limitations, we created a framework that enables us to integrate reconstructions of different state-of-the-art SfM and MVS libraries into Blender - see \figref{fig_imported_result} for an example. Since the source code of all components is publicly available, the full pipeline is not only suitable for modeling or creating visual effects, but especially for future research efforts. \\
At the same time, our framework servers in combination with Blender's animation and rendering capabilities as a tool for the photogrammetry community that offers sophisticated inspection and visualization functionalities, which are not present in other editors such as Meshlab \citep{Cignoni2008Meshlab} or CloudCompare \citep{CloudCompare2020}.


\subsection{Related Work}

SfM is a photogrammetric technique that estimates for a given set of (unordered) input images the corresponding three-dimensional camera poses and scene structures. There are two categories of SfM approaches: \emph{incremental} and \emph{global} SfM. Incremental SfM is currently the prevalent state-of-the-art method \citep{Schoenberger2016sfm}. In order to manage the problem complexity of reconstructing real-world scenes, incremental SfM decomposes the reconstruction process into more controllable subproblems. The corresponding tasks can be categorized in \emph{correspondence search} (including \emph{feature detection}, \emph{feature matching} and \emph{geometric verification}) and \emph{sparse reconstruction} (consisting of \emph{image registration}, \emph{point triangulation}, \emph{bundle adjustment} and \emph{outlier filtering}). During bundle adjustment, SfM minimize the reprojection error of the reconstructed three-dimensional points for each view. \\
MVS uses the camera poses and the sparse point cloud obtained in the SfM step to compute a dense point cloud or a (textured) model reflecting the geometry of the input scene. Similarly to SfM, MVS divides the reconstruction task in multiple subproblems. The \emph{multi-view stereo} step computes a depth map for each registered image that potentially includes surface normal vectors. \emph{Multi-view fusion} fuses the depth maps into a unified dense reconstruction that allows to reconstruct a watertight surface model in the \emph{surface reconstruction} step. \figref{figure_structure_from_motion_pipeline} shows an overview of essential SfM and MVS subtasks and their dependencies. \\
Currently, there are several state-of-the-art photogrammetry libraries that provide full SfM and MVS pipelines such as Colmap \citep{Colmap2020},  Meshroom \citep{Meshroom2018}, Multi-View Environment \citep{MVE}, OpenMVG \citep{OpenMVG} \& OpenMVS \citep{OpenMVS2020} as well as Regard3D \citep{Regard3D2020}. For a quantitative evaluation of state-of-the-art SfM and MVS pipelines on outdoor and indoor scenes see \cite{Knapitsch2017}, which provides a benchmark dataset using laser scans as ground truth. \\
While the usage of the reconstructed (textured) models is widely supported by modern modeling tools, the integration of camera-specific information such as intrinsic and extrinsic parameters are oftentimes neglected. There are only a few software packages available that allow to import camera-specific information into modeling programs. The majority of these packages address specific proprietary reconstruction or modeling tools such as \cite{MeshroomMaya2020}, \cite{GameDevelopmentToolset2020} or \cite{PhotoscanBlender2020}. The most similar tool compared to the proposed framework is presumably \cite{BlenderPhotogrammetry2020}, which also provides options to import SfM and MVS formats into Blender. However, the following capabilities of our framework are missing in \cite{BlenderPhotogrammetry2020}: visualization of colored point clouds, representation of source images as image planes and creation of point clouds from depth maps. Further, \cite{BlenderPhotogrammetry2020} supports less SfM and MVS libraries and provides less options to configure the input data.

\begin{table*}[t]
	\centering
	\begin{tabular}{c c c } 
		\textbf{Pipeline} 		& \textbf{Colmap}	 			& \textbf{Meshroom}  \\
		Structure from Motion 					& \cite{Schoenberger2016sfm} 	& \cite{Moulon2012ACCV}\\
		Multi-view Stereo 					& \cite{Schoenberger2016mvs}	& \cite{HirschmullerCVPR2005} \\ 
		Mesh Reconstruction  	& \cite{Kazhdan2013}			& \cite{JancosekISRN2014} \\
		Texturing 				& -  				 			& \cite{BurtToG1983}\\
								&								&	\\
		\textbf{Pipeline}		& \textbf{MVE}					& \textbf{OpenMVG / OpenMVS} \\
		Structure from Motion 					& \cite{MVE} 					& \cite{Moulon2012ACCV} \\
		Multi-view Stereo 					& \cite{GoeseleICCV2007} 		& \cite{BarnesToG2009}  \\ 
		Mesh Reconstruction  	& \cite{FuhrmannToG2014} 		& \cite{JancosekISRN2014} \\
		Texturing 				& \cite{WaechterECCV2014} 		& \cite{WaechterECCV2014}  \\
								&								&	\\
		\textbf{Pipeline}		& \textbf{Regard3D}				& \textbf{VisualSfM} \\
		Structure from Motion 					& \cite{Moulon2012ACCV}			& \cite{Moulon2012ACCV} \\
		Multi-view Stereo 					& \cite{LangguthECCV2016SMVS}	& \cite{FuruTPAMI2010PMVS} \\
		Mesh Reconstruction  	& \cite{FuhrmannToG2014}		& - \\
		Texturing 				& \cite{WaechterECCV2014}		& - \\
	\end{tabular}
	\caption{Overview of photogrammetry pipelines that are supported by the proposed framework. In many cases the pipelines allow to substitute specific pipeline steps using alternative implementations. This table shows the default or the recommended pipeline configurations.}
	\label{table_sota_rec_pipelines}
\end{table*}

\subsection{Contribution}

The core contribution of this work are as follows. 
(1)~The proposed framework allows to leverage image-based reconstructions (\eg automatic calibration of intrinsic camera parameters, computation of three-dimensional camera poses and reconstruction of scene structures) for different tasks in Blender such as sculpting or creating visual effects.\\
(2)~We use available data structures in Blender to represent the integrated reconstruction results, which allows the framework to compute automatic camera animations (including extrinsic and intrinsic camera parameters), represent the reconstructed point clouds as particle system or attach the source images to the registered camera poses. Using the available data structures in Blender ensures that the integrated results can be further utilized.\\
(3)~This framework provides (together with Blender's built-in tools) different visualization and animation capabilities for image-based reconstructions that are superior to tools offered by common photogrammetry-specific software packages. \\
(4)~The framework supports already many state-of-the-art open source SfM and MVS pipelines and is (because of its modular design) easily extensible. \\
(5)~The source code of framework is publicly available\footnote{\label{source_code}Source code is available at \url{https://github.com/SBCV/Blender-Addon-Photogrammetry-Importer}}.

\begin{figure*}[t]
	\centering
	{
		\begin{subfigure}[t]{0.49\textwidth}
			\includegraphics[width=\textwidth]{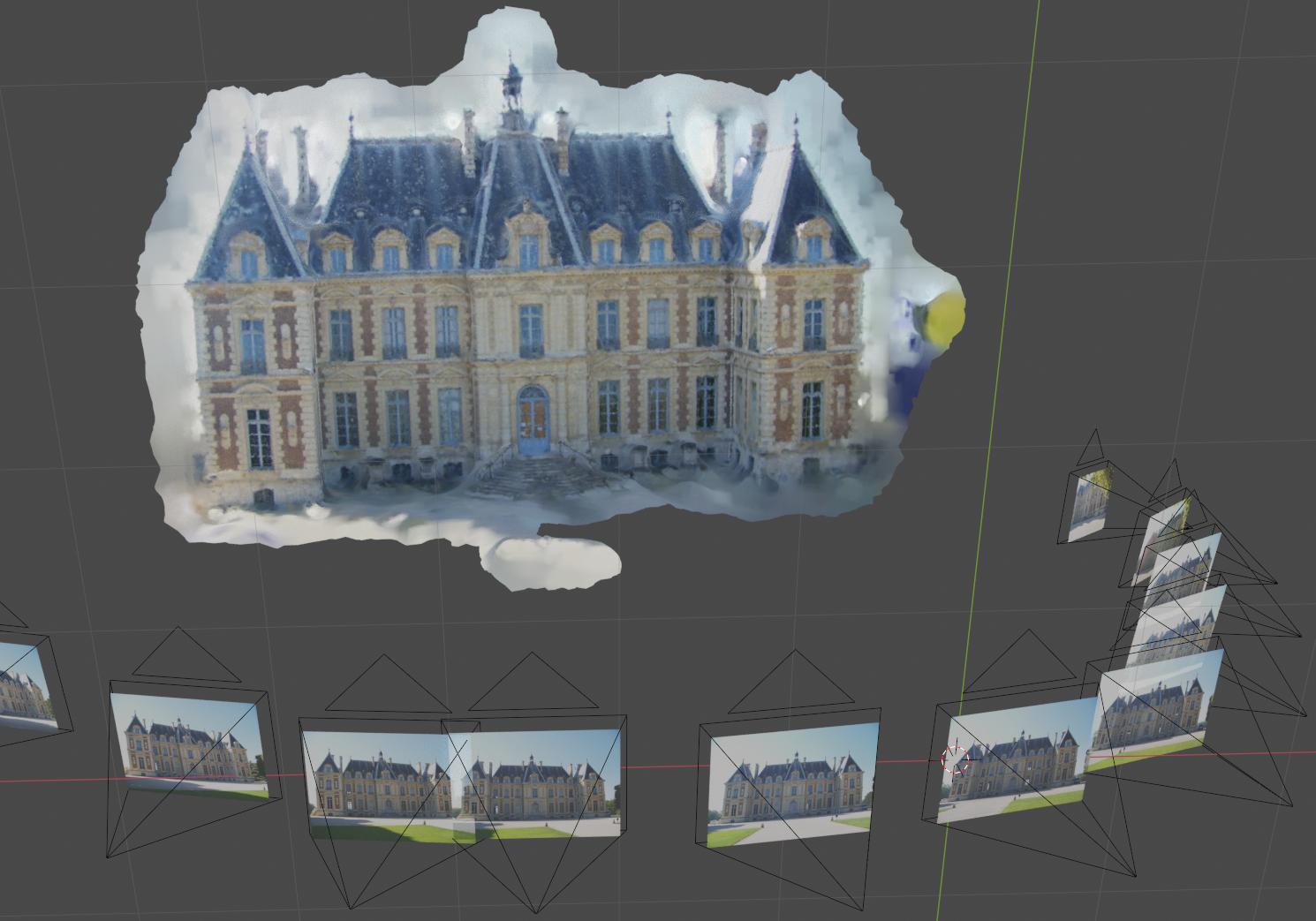}
			\caption{Reconstructed geometry represented as mesh with vertex colors in the 3D view of Blender.}
			\label{subfig_mesh_in_3d_view}
		\end{subfigure}
		\hfill
		\begin{subfigure}[t]{0.49\textwidth}
			\includegraphics[width=\textwidth]{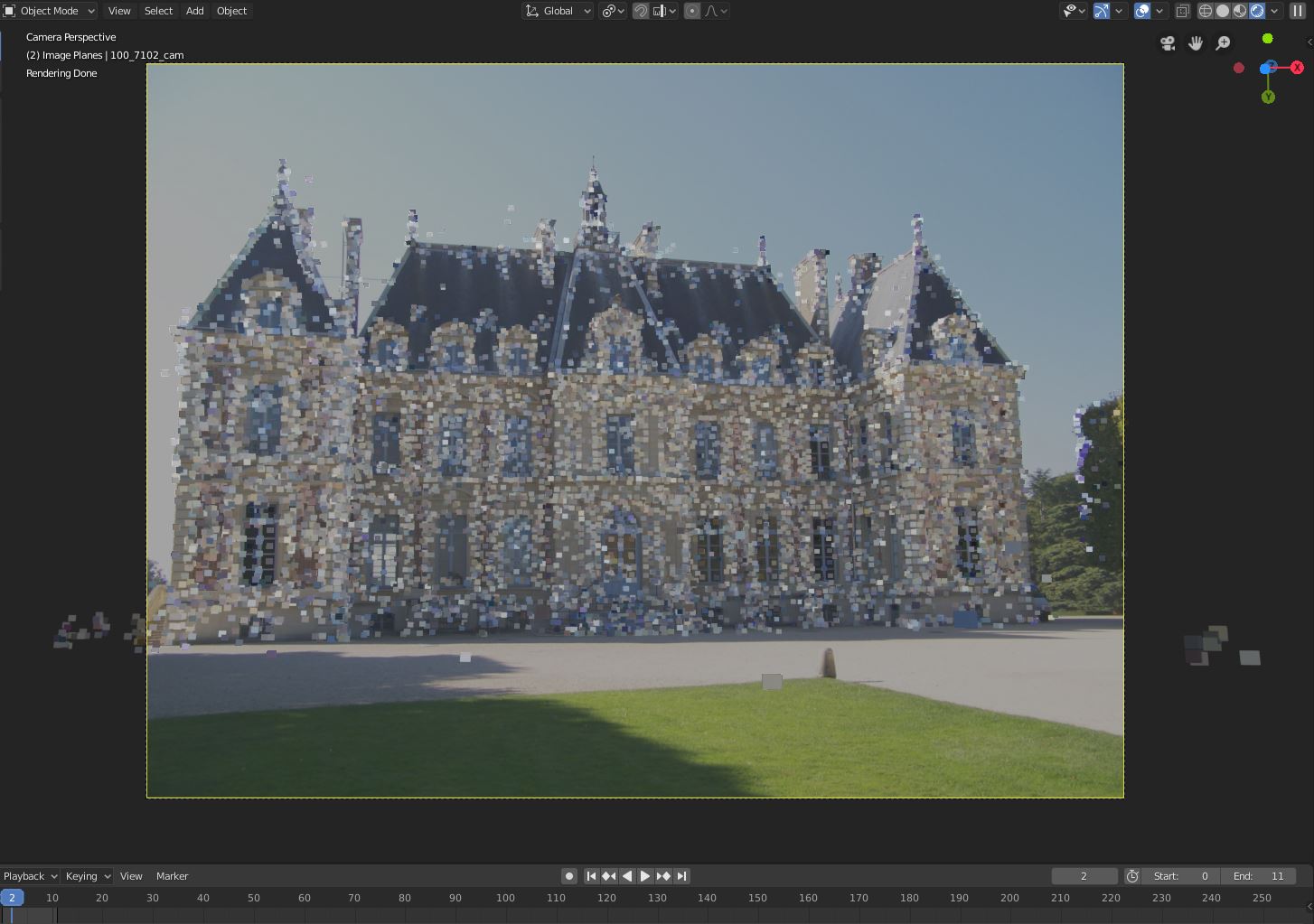}
			\caption{Reconstructed point cloud in the 3D view of Blender from the perspective of one of the reconstructed cameras. The background image shows the corresponding input image used to register the camera.}
			\label{subfig_background_image}
		\end{subfigure}	
	}
	\caption{Reconstructed geometry in the 3D view of Blender.}
	\label{fig_geometry_in_3d_view}
\end{figure*}

\section{\uppercase{Framework}}

\subsection{Overview}

\noindent The proposed framework allows us to import the reconstructed scene geometry represented as point cloud or as (textured) mesh, the reconstructed cameras (including intrinsic and extrinsic parameters), point clouds corresponding to the depth maps of the cameras and an animated camera representing the camera motion. The supported libraries include the following photogrammetry pipelines: Colmap \citep{Colmap2020}, Multi-View Environment \citep{MVE}, OpenMVG \citep{OpenMVG} \& OpenMVS \citep{OpenMVS2020} and Meshroom \citep{Meshroom2018} as well as VisualSfM \citep{Wu2011}. \tabref{table_sota_rec_pipelines} contains an overview of each pipeline with the corresponding reconstruction steps. An example reconstruction result of Colmap is shown in \figref{subfig_mesh_in_3d_view}. \\
In addition to the SfM and MVS libraries mentioned above, the system supports the integration of camera poses or scene structures captured with RGB-D sensors using \cite{Zhou2018} as well as point clouds provided in common laser scanning data formats.
\\
We followed a modular design approach in order to simplify the extensibility of the framework. Each supported library requires the implementation of a \emph{file handler} that is able to read the corresponding file format and an \emph{import operator} that defines library specific options for importing the corresponding results.

\begin{figure*}[t]
	\centering
	{
		\begin{subfigure}[t]{0.44\textwidth}
			\includegraphics[width=\textwidth]{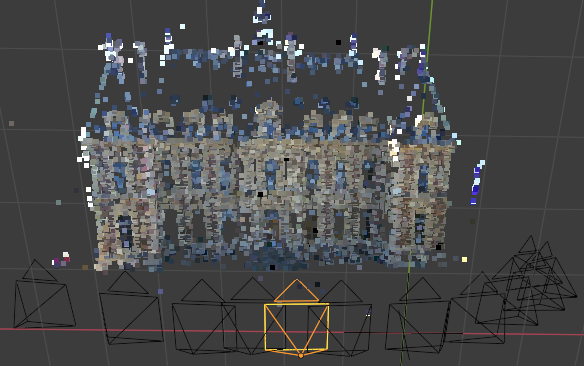}
			\caption{Reconstruction result in Blender's 3D view. The reconstructed cameras are shown in black and the animated camera in orange, respectively.}
		\end{subfigure}
		\hfill
		\begin{subfigure}[t]{0.55\textwidth}
			\includegraphics[width=\textwidth]{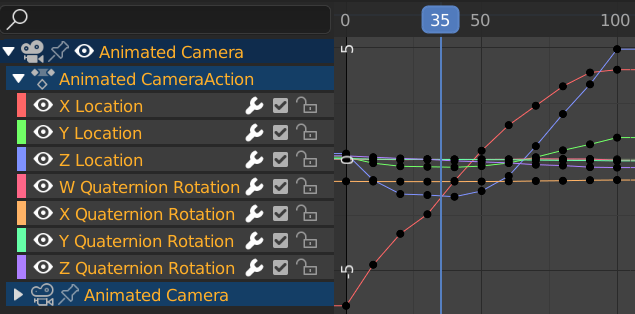}
			\caption{Interpolation values of the translation and the rotation corresponding to the animated camera in the left image. The black dots denote the values of the reconstructed camera poses and the vertical blue line indicates the interpolated values at the position of the camera in the left image.}
		\end{subfigure}
		
	}
	\caption{Example of a camera animation using 11 images of the Sceaux Castle dataset \citep{Moulon2012Sceaux}. By interpolating the poses of the reconstructed cameras, we obtain a smooth trajectory for the animated camera.}
	\label{fig_f_curves}
\end{figure*}

\begin{figure*}[t]
	\centering
	{
		\includegraphics[ width=\textwidth, trim={0 15cm 0 0},clip]{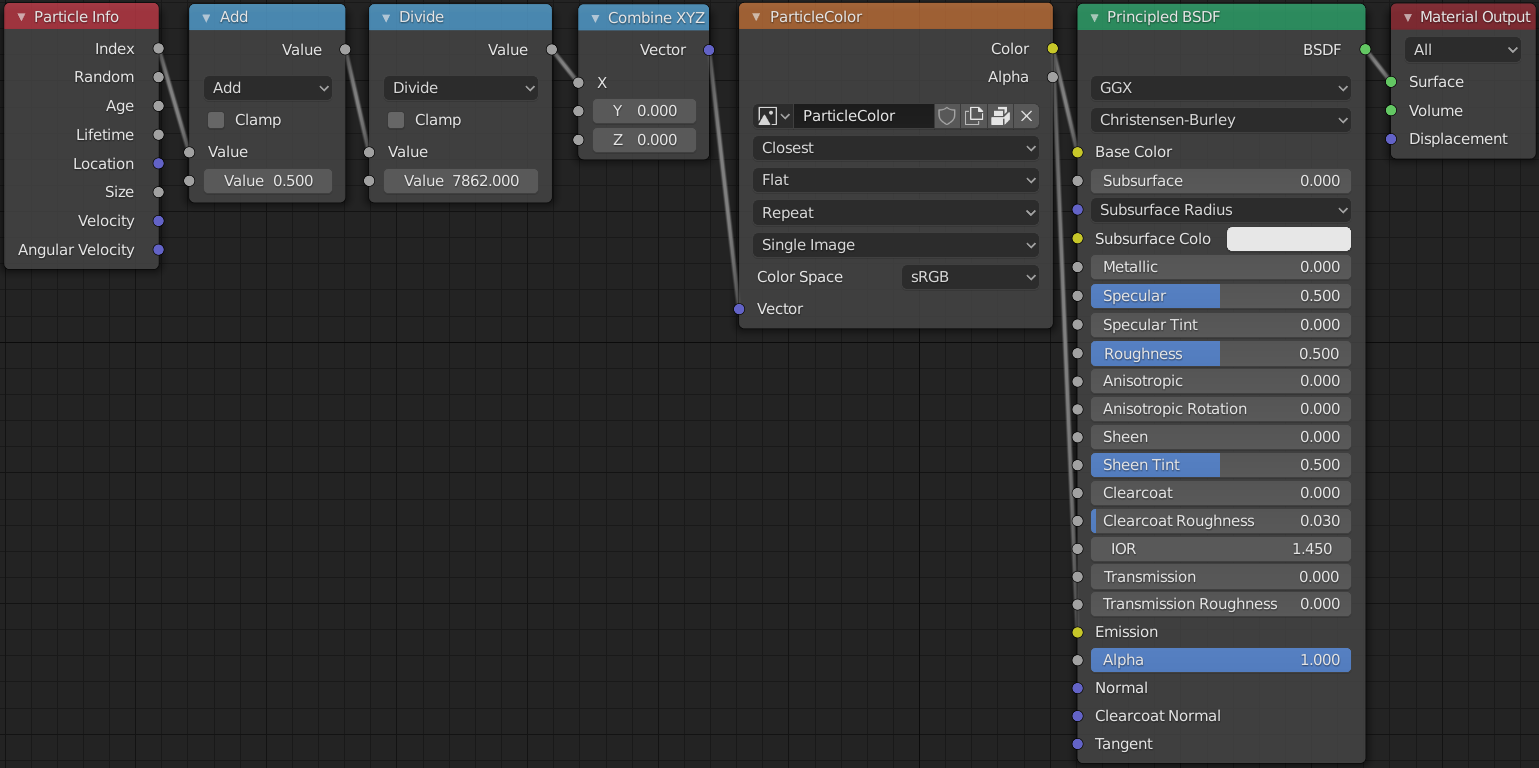}
	}
	\caption{Node configuration created by the framework to define colors of the particles in the particle system. The light blue nodes use the particle index to compute the texture coordinate with the corresponding color - the value in the divide node represents the number of total particles. The principled BSDF node has been cut to increase the compactness of the figure.}
	\label{fig_particle_system_material}
\end{figure*}
\begin{figure*}[t]
	\centering
	{
		\begin{subfigure}[t]{0.49\textwidth}
			\includegraphics[width=\textwidth]{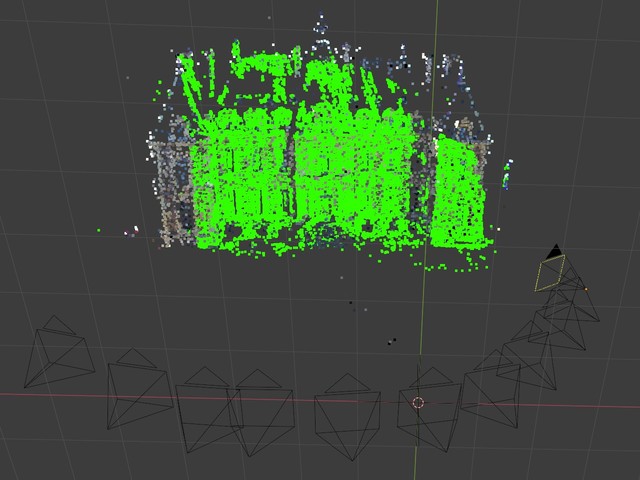}
			\caption{Reconstructed point cloud and depth map of the rightmost camera highlighted with yellow.}
			\label{subfig_depth_map_visualization_3d_view}
		\end{subfigure}
		\hfill
		\begin{subfigure}[t]{0.49\textwidth}
			\includegraphics[width=\textwidth]{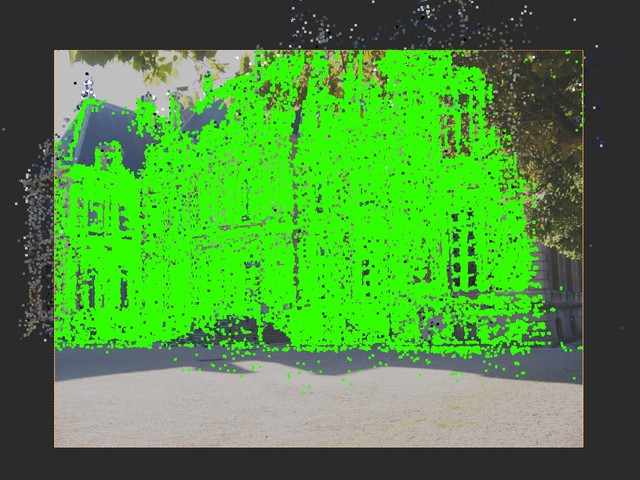}
			\caption{Depth map from the perspective of one of the reconstructed cameras. The background image shows the input image used to register the camera.}
			\label{subfig_depth_map_visualization_from_camera}
		\end{subfigure}
	}
	\caption{Representation of a depth map using OpenGL. Both images show the triangulated points corresponding to the depth map of the same camera. Using Blender's 3D view allows us to assess the consistency of the depth map \wrt to the point cloud (see \figref{subfig_depth_map_visualization_3d_view}) as well as the consistency of the depth \wrt to visual cues in source image (\figref{subfig_depth_map_visualization_from_camera}).}
	\label{fig_depth_map_visualization}
\end{figure*}

\subsection{Camera and Image Representation}

\noindent The supported SfM and MVS libraries use different camera models to represent the intrinsic camera parameters during the reconstructions process. Further, the conventions describing the extrinsic camera parameters are also inconsistent. We convert the different formats into a unified representation that can be directly mapped to Blender's \emph{camera objects}. \\
In addition to the integration of the geometric properties of the reconstructed cameras, the framework provides an option to add each input image as \emph{background image} for the corresponding camera object. Viewing the scene from the perspective of a specific camera allows to assess the consistency of virtual objects and the corresponding source images, which is especially useful for sculpting tasks. It also offers convenient capabilities to visualize and inspect the reconstructed point clouds and meshes, which are not feasible with other photogrammetry-specific tools such as CloudCompare and Meshlab. \figref{subfig_background_image} shows for example a comparison of the projected point cloud and the color information of the corresponding input image. \\
To further enhance the visualization, the system provides an option to add the original input images as separate image planes as shown in \figref{fig_imported_result}. \\
In order to easy the usage of the reconstruction for animation and visual effect tasks, the framework offers an option to create an animated camera using the reconstructed camera poses as well as the corresponding intrinsic parameters such as the focal length and the principal point. All parameters are animated by using Blender's built-in \emph{f-curves}, which allows to post-process the result with Blender's animation tools. The camera properties between two reconstructed camera poses are interpolated to enable the creation of smooth camera trajectories. \figref{fig_f_curves} shows an example of the animated camera and the corresponding interpolated properties. The f-curves use quaternions to define the camera rotations of the animated camera. We normalize the quaternions representing the roations of the reconstructed cameras to avoid unexpected interpolation results caused by quaternions with different signs, \ie we ensure that the quaternions show consistent signs, which minimizes the distance between two consecutive quaternions.

\subsection{Representation of Scene Geometry}

\noindent Photogrammetry-based reconstruction techniques frequently use two types of entities to represent the reconstructed scene structure: point clouds and (textured) meshes. While meshes provide a more holistic representation of the scene geometry, point clouds are typically more accurate - since the reconstructed points correspond to image correspondences. \\
Currently, there is no Blender entity that permits to directly represent colored point clouds. Our frameworks circumvents this problem by providing the following two point cloud representations: point clouds represented with Blender's \emph{particle system} and point clouds visualized with \emph{OpengGL} \citep{Woo1999opengl}. \\
The particle system allows us to represent each 3D point with a single particle, which enables us to post-process and render the reconstructed result. We define the particle colors with Blender's \emph{node system}. The proposed framework uses the ID of each particle to determine a texture coordinate of single texture containing all particle colors. The corresponding nodes are shown in \figref{fig_particle_system_material}. \\
In contrast to Blender's particle system, the drawing of point clouds with OpenGL is computationally less expensive and enables to visualize larger point clouds. Thus, it is better suited for the visualization of large point numbers such as point clouds representing the depth maps of multiple input images. \figref{fig_depth_map_visualization} shows an example. \\
We use Blender's built-in data structures to represent the reconstructed and potentially textured meshes. This allows us to integrate the triangulated scene geometry (\ie points and meshes) in the same coordinate system as the registered cameras as shown in \figref{fig_geometry_in_3d_view}.



\section{\uppercase{Conclusion}}
\noindent This paper proposes a publicly available extension of Blender that enables to integrate reconstructions of different SfM and MVS libraries to facilitate tasks such as image-based modeling and automatic creation of camera animations. We presented a general overview of the core components of modern SfM and MVS methods and provided a summary of widely used state-of-the-art reconstruction pipelines that are supported by our system. The paper provides a detailed description of the objects used to model the reconstructed cameras and the corresponding three-dimensional scene points as well as the representations used to integrate camera motion and additional image information. We showed several automatically computed examples that illustrate the usefulness of the presented framework. We are convinced that the application of SfM and MVS libraries is a crucial step to facilitate image-based modeling and camera tracking tasks. The analysis of the download statistics show that our Blender extension is adopted by a variety of artists and researchers.

%
%
%

\bibliographystyle{apalike}
{\small
\bibliography{mybib}}

\end{document}